\documentclass[pdflatex,sn-mathphys-num]{sn-jnl}

\usepackage{graphicx}%
\usepackage{multirow}%
\usepackage{amsmath,amssymb,amsfonts}%
\usepackage{amsthm}%
\usepackage{mathrsfs}%
\usepackage[title]{appendix}%
\usepackage{xcolor}%
\usepackage{textcomp}%
\usepackage{manyfoot}%
\usepackage{booktabs}%
\usepackage{algorithm}%
\usepackage{algorithmicx}%
\usepackage{algpseudocode}%
\usepackage{listings}%

\theoremstyle{plain}%
 
\theoremstyle{plain}%

\theoremstyle{plain}%

\usepackage{subcaption}
\usepackage{array}

\newcommand{\numparties}{\mathcal{N}}

\newcommand{\pone}{p} 
\newcommand{\ptwo}{q} 

\newcommand{\fea}{f} 

\newcommand{\cd}{D} 

\begin{document}

\title[Accurate and Private Diagnosis of Rare Genetic Syndromes from Facial Images with Federated Deep Learning]{Accurate and Private Diagnosis of Rare Genetic Syndromes from Facial Images with Federated Deep Learning}


\author*[1,2]{\fnm{Ali Burak} \sur{Ünal}}\email{ali-burak.uenal@uni-tuebingen.de}

\author[1,2]{\fnm{Cem Ata} \sur{Baykara}}\email{cem.baykara@uni-tuebingen.de}

\author[3]{\fnm{Peter} \sur{Krawitz}}\email{pkrawitz@uni-bonn.de}


\author[1,2]{\fnm{Mete} \sur{Akgün}}\email{mete.akguen@uni-tuebingen.de}

\affil*[1]{\orgdiv{Medical Data Privacy and Privacy Preserving Machine Learning (MDPPML)}, \orgname{University of Tübingen}, \orgaddress{\city{Tübingen}, \postcode{72076}, \country{Germany}}}

\affil[2]{\orgdiv{Institute for Bioinformatics and Medical Informatics (IBMI)}, \orgname{University of Tübingen}, \orgaddress{\city{Tübingen}, \postcode{72076}, \country{Germany}}}

\affil[3]{\orgdiv{Institute for Genomic Statistics and Bioinformatics}, \orgname{University of Bonn}, \orgaddress{\city{Bonn}, \postcode{53113}, \country{Germany}}}


\abstract{

Machine learning has shown promise in facial dysmorphology, where characteristic facial features provide diagnostic clues for rare genetic disorders. GestaltMatcher, a leading framework in this field, has demonstrated clinical utility across multiple studies, but its reliance on centralized datasets limits further development, as patient data are siloed across institutions and subject to strict privacy regulations. We introduce a federated GestaltMatcher service based on a cross-silo horizontal federated learning framework, which allows hospitals to collaboratively train a global ensemble feature extractor without sharing patient images. Patient data are mapped into a shared latent space, and a privacy-preserving kernel matrix computation framework enables syndrome inference and discovery while safeguarding confidentiality. New participants can directly benefit from and contribute to the system by adopting the global feature extractor and kernel configuration from previous training rounds. Experiments show that the federated service retains over 90\% of centralized performance and remains robust to both varying silo numbers and heterogeneous data distributions.}




\maketitle

\section{Introduction} \label{sec:intro}
Rare diseases, often defined as conditions affecting a small percentage of the population, pose significant challenges to the medical community. Despite their rarity, the cumulative impact of these diseases is substantial, affecting millions of individuals worldwide \cite{schieppati2008rare,ferreira2019burden}. The complexity and diversity of rare diseases necessitate comprehensive databases to improve understanding, diagnosis, and treatment \cite{liu2022natural}.
One of the primary challenges in diagnosing and understanding rare diseases is the scarcity and fragmentation of patient records \cite{dos2022towards,banerjee2023machine}. Given the small number of cases, relevant data is often dispersed across various institutions and regions, making it difficult to gather a sufficient dataset for robust analysis. This dispersion complicates efforts to identify common patterns and develop effective diagnostic tools.

To overcome these challenges, initiatives such as GestaltMatcher have been developed \cite{hsieh2022gestaltmatcher}. In GestaltMatcher services, a comprehensive image database called GestaltMatcher database (GMDB) and predictive/discovery services are provided. GMDB aims to aggregate imaging data of individuals with rare dysmorphic features, providing a comprehensive reference set for clinicians and a training resource for computer scientists developing medical artificial intelligence systems. By bringing together different data sources, GMDB enhances the potential for accurate diagnosis and the discovery of new syndromes.
Although comprehensive data sets on patients with rare disorders are extremely valuable for research and diagnostics, sharing such data with centralized platforms is often prohibited legally. Even though GDPR \cite{gdpr} was established to enable the secure and free flow of personal data within the European Union, it links this to two essential conditions: the explicit consent of the individual concerned or a compelling need for medical care. If these requirements are not met, the data cannot be legally transferred, regardless of whether the data custodians would otherwise be willing to collaborate. 

Considering its ability to distribute computation between multiple sources and eliminate the need for centralized data collection for training and inference purposes, federated learning \cite{konevcny2016federated,mcmahan2017communication} is a natural candidate to address the privacy issues inherent in the GestaltMatcher database. 
For federated learning to be successfully applied to a problem, the data distribution must first be analyzed.
In \textit{cross-siloed} federated learning, the participants act as ``silos'' and want to collaborate with other \textit{silo} participants to have a better perspective on and solution to the problem. In \textit{cross-device} federated learning, on the other hand, the central server desires to have a central model trained on the data of, generally, edge devices without seeing their data. 
Another aspect of differentiating federated learning solutions is based on how data is distributed between participants \cite{yang2019federated}. If the sample sets are distributed across participants, these systems are called \textit{horizontal federated learning}. If the feature sets of the same samples are distributed among participants, then such federated systems are called \textit{vertical federated learning}.
This flexibility of federated learning makes it even more impactful and attractive, and it is widely utilized to address privacy and computational concerns in many different problems such as drug discovery \cite{chen2020fl}, face recognition \cite{niu2022federated}, end-to-end private medical imaging \cite{kaissis2021end,kaissis2020secure}, cancer boundary detection \cite{pati2022federated}, disease module discovery and classification \cite{pfeifer2023ensemble}, identifying high-risk patients with postoperative gastric cancer recurrence \cite{feng2024robustly}, and differentially private medical image analysis \cite{adnan2022federated}. In our case, the scenario primarily pertains to \textit{cross-siloed horizontal} federated learning, as we are working with patient data distributed across different hospitals and a patient is typically associated with only one hospital.

To address data owners’ privacy concerns and advance research on syndromes with facial dysmorphism, we propose a cross-siloed horizontal federated GestaltMatcher service. This decentralized approach allows data owners to contribute to collaborative efforts to understand these syndromes without needing to share patient images with a central server. They not only have the distributed collective database allowed by the privately trained global feature extractor but also benefit from the prediction and discovery opportunities, which is otherwise not possible. Our federated GestaltMatcher service provides robust predictive capabilities and facilitates the discovery of ultra-rare syndromes while ensuring that patient images remain private. Rather than transmitting images or their plaintext latent representations, data owners share only masked versions of their image encodings, enabled by a privacy preserving kernel matrix computation framework. This allows data owners to benefit from GestaltMatcher service’s predictive power by inferring the syndrome of a query patient and its capacity to identify highly similar patient subgroups, which may prompt coordinated studies and potentially reveal novel disease phenotypes without compromising privacy. We extensively analyzed the performance of our method on the GestaltMatcher dataset for different parameters of the system, validating its performance while keeping the data private. In summary, our contributions can be listed as follows:
\begin{itemize}
    \item We introduce a federated decentralized GestaltMatcher service to enable data owners with privacy concerns to contribute to the collaborative data collection effort of understanding Gestalt syndromes.

    \item We provide data owners with privacy preserving collaborative training of a global ensemble feature extractor model via federated learning that map the samples distributed across multiple sources to the same latent space, forming a distributed GestaltMatcher database.

    \item We enable data owners to benefit from the predictive and exploratory power of the service by employing a privacy preserving kernel matrix computation framework, called FLAKE \cite{hannemann2023privacy}, on the distributed latent representation of patients' facial images to privately compute the pairwise distances, allowing to perform inference of patient's syndrome and identify exceptionally similar small patient groups that could lead to a new syndrome.

    \item We keep the predictive power of the centralized ensemble model almost the same in our federated GestaltMatcher service, having at least $90\%$ of the centralized version's performance for any setting in our experiments.

    \item We systematically evaluate the scalability and robustness of the federated GestaltMatcher service by varying the number of participating silos and the degree of data heterogeneity across clients. Our results demonstrate that the model maintains stable performance and consistent feature representations even under highly imbalanced and non-IID data distributions.
\end{itemize}

\section{GestaltMatcher Services} \label{sec:gm}
\begin{figure}
    \centering
    \includegraphics[width=\linewidth]{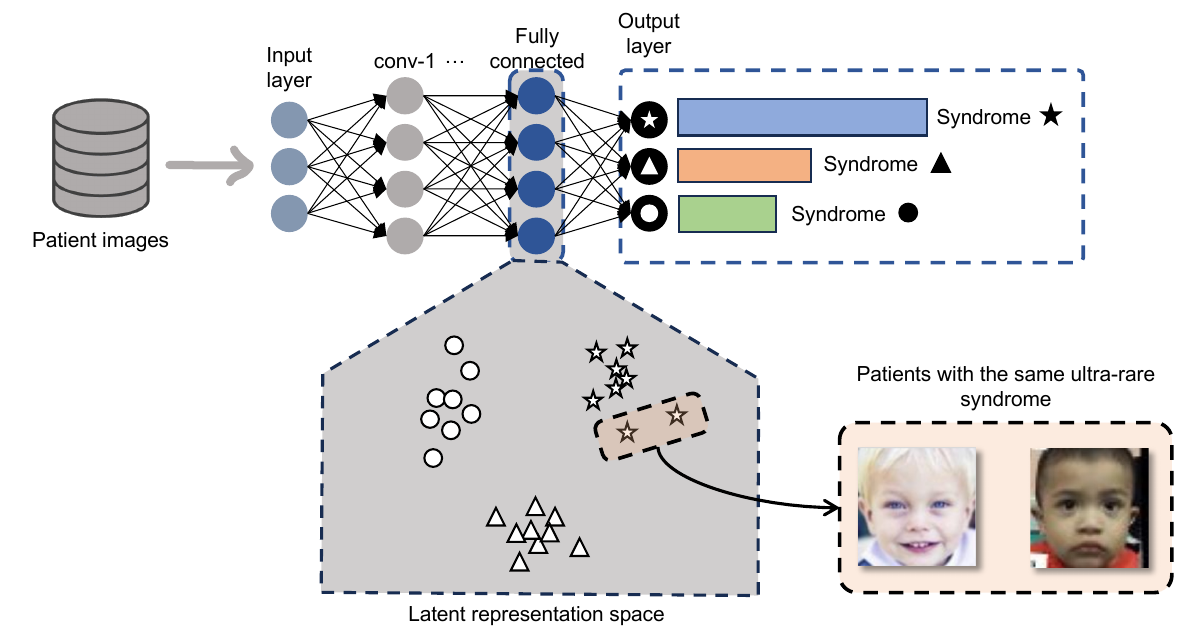}
    \caption{Centralized GestaltMatcher service. A deep neural network is trained on patient images to extract facial features and map them into a latent representation space. Patients with similar features cluster together in this space, enabling syndrome diagnosis and the identification of ultra-rare disorders.}
    \label{fig:central_gm}
\end{figure}

GestaltMatcher represents a suite of next-generation phenotyping tools designed to leverage artificial intelligence and data-sharing frameworks aligned with the principles of findability, accessibility, interoperability, and reusability (FAIR) \cite{wilkinson2016fair} to assist in the diagnosis of rare genetic disorders. By employing advanced computer vision techniques such as deep convolutional neural networks, GestaltMatcher identifies characteristic dysmorphic features from medical imaging data (Figure~\ref{fig:central_gm}). These features enable the support of clinical workflows, aiding in dysmorphism assessment and the classification of genetic variants.

The GestaltMatcher services consist of algorithms and a centralized database called GMDB. GMDB is a curated repository containing clinical features annotated with human phenotype ontology terminology, molecular diagnoses, and medical imaging data. It provides a foundation for training and evaluating the algorithms, which are designed to match patients in a high-dimensional feature space, facilitating the diagnosis of both common, rare and ultra-rare genetic disorders. The centralized nature of GMDB allows for the consolidation of data, enabling the development of highly accurate models. For example, \citet{hustinx2023improving} demonstrated significant advancements in facial phenotyping by introducing updated architectures and ensemble modeling techniques. They replaced outdated components of GestaltMatcher, such as the earlier AlexNet-based architecture, with state-of-the-art iResNet and ArcFace methods, which improved the accuracy of disorder recognition. Additionally, they employed test-time augmentation and model ensembles to optimize the system’s performance on both seen and unseen syndromes. This work established a new benchmark by achieving superior results in verifying ultra-rare disorders, which are often the most challenging to diagnose due to limited data.

Despite these advances, the centralized approach of GestaltMatcher poses several challenges. Privacy concerns discourage individuals and hospitals from contributing their data, reducing the diversity and volume of data available in GMDB. This limitation can lead to suboptimal model performance and biases in the representations, as the data may not adequately cover all populations or rare conditions. Furthermore, some institutions may be unable or unwilling to use the centralized diagnostic service due to regulatory or institutional privacy constraints, potentially depriving patients of accurate and timely diagnoses. When institutions attempt to fine-tune or locally use the GestaltMatcher model on their own data, they often end up with inferior model performance compared to a globally trained model. This fragmentation not only reduces diagnostic accuracy but also diminishes the potential to identify new syndromes emerging within localized datasets.

\section{Results} \label{sec:results}
In this section, we will provide a brief introduction to the federated GestaltMatcher service, followed by a detailed description of the dataset used in our experiments. Before presenting the experimental results that evaluate our solution based on various criteria, we will explain the experimental setup.

\subsection{Overview of Federated GestaltMatcher} \label{sec:overview_fedgm}
To provide a decentralized GestaltMatcher database and services, we designed a federated GestaltMatcher service. In this solution, data owners such as hospitals, healthcare centers, and institutions do not require to sacrifice the privacy of their data to contribute the collaborative effort to better understand and diagnose Gestalt syndromes. 

\begin{figure}
    \centering
    \includegraphics[width=\linewidth]{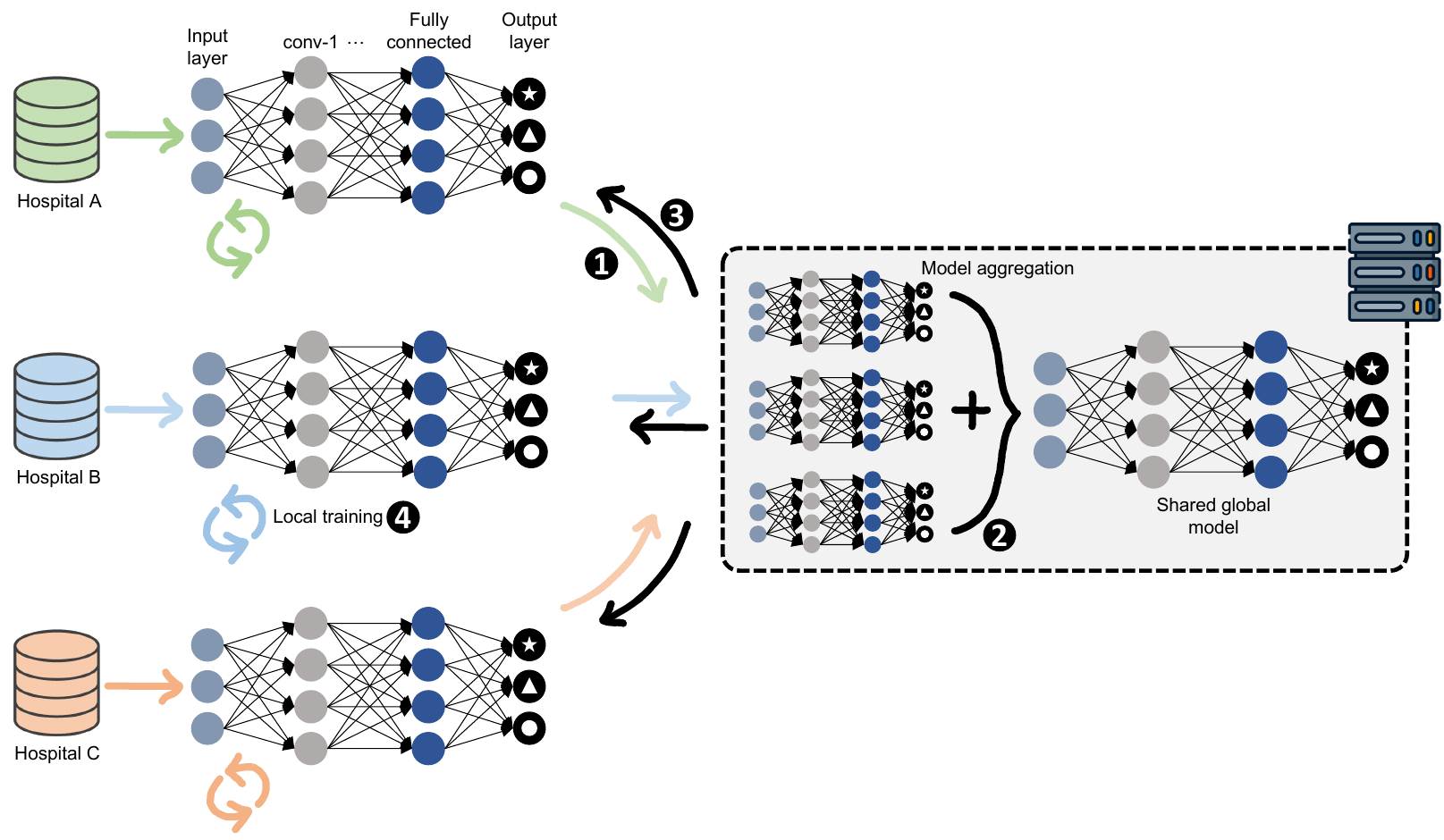}
    \caption{Federated training of the global ensemble feature extraction model. \textbf{(1)} The local models are first shared with the aggregator. \textbf{(2)} The aggregator aggregates these local models and \textbf{(3)} sends it to the silos. \textbf{(4)} The silos continue training the aggregated model using their local data.}
    \label{fig:first_phase_overview}
\end{figure}

In this distributed solution, we use the same ensemble model proposed by \citet{hustinx2023improving} without requiring to form a central database of patient images. Data owners train this ensemble feature extractor model using federated learning as illustrated in Figure \ref{fig:first_phase_overview}. All data owners start with the same initial models. They locally train these initial models for some epochs using their local data and then send them to a central server, which we call \textit{aggregator}, to aggregate these local models and obtain global intermediate models. We employ secure aggregation in this process to keep the local models and the global intermediate models private. Once the global intermediate models are calculated, the aggregator sends them back to the data owners for further training. This cycle of training and aggregating of local models continues for a predetermined number of rounds until the data owners obtain the final global models of the ensemble feature extractor.

\begin{figure}[!th]
    \centering
    \includegraphics[width=\linewidth]{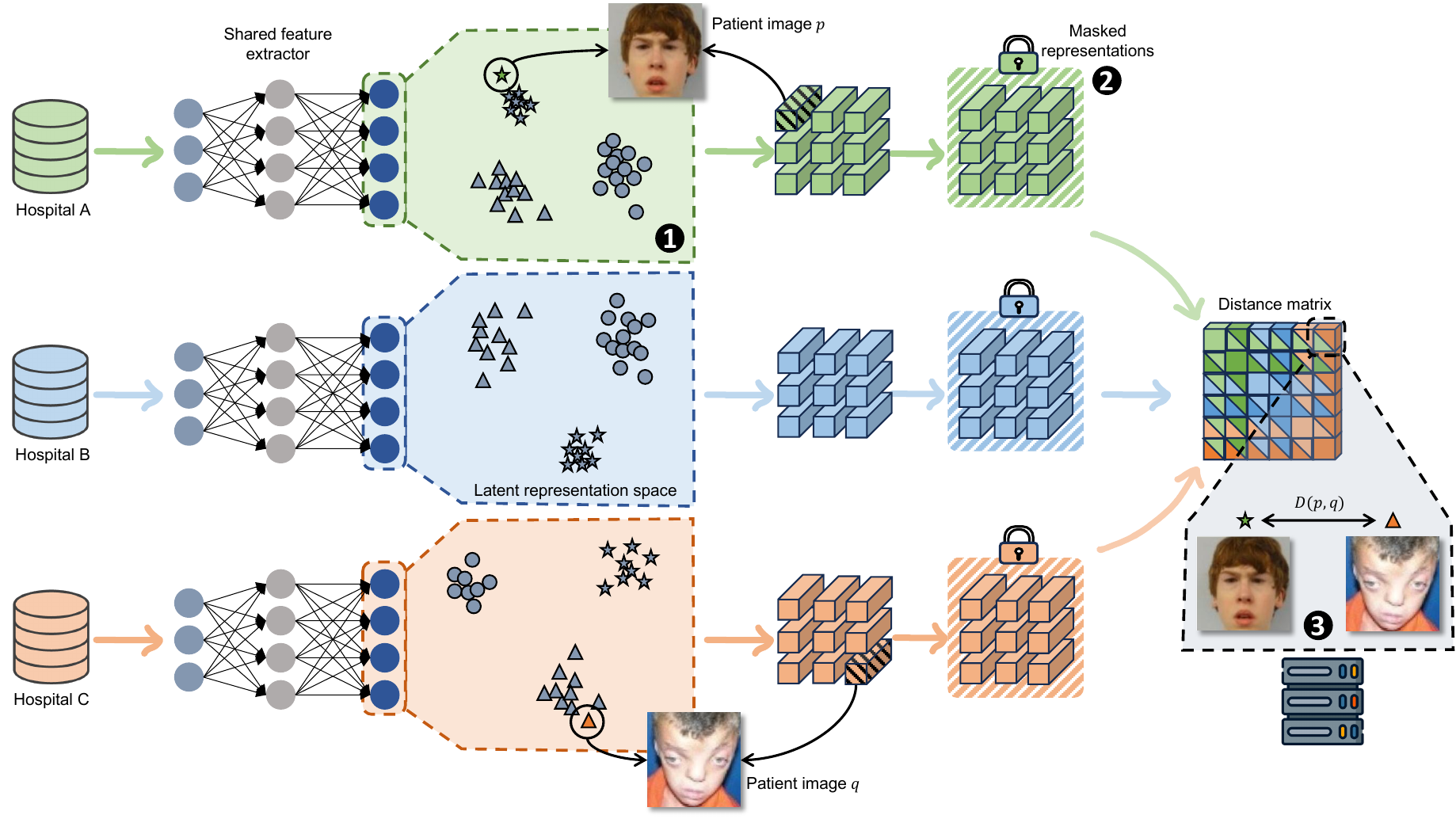}
    \caption{\textbf{(1)} The silos extract the features of their local data using the final aggregated model. \textbf{(2)}  They mask the latent representations of their images and send them to the aggregator. \textbf{(3)} The aggregator computes the cosine distance matrix from the masked latent representations.}
    \label{fig:cosine_distance_matrix_computation}
\end{figure}

Using the collaboratively and privately trained global ensemble feature extractor model, the data owners compute the latent representations of their patient images so that their distances in this space can be computed for inference and discovery services. To handle this distance computation privately, we employ a special privacy preserving kernel matrix computation framework, called FLAKE \cite{hannemann2023privacy}. It allows data owners to compute all pairwise dot products of their images' latent representations in the aggregator, later used to compute the cosine distances of these images' latent representations without sharing the images or their representations. This process is depicted in Figure \ref{fig:cosine_distance_matrix_computation}.




\begin{figure}[!th]
    \centering
    \includegraphics[width=0.9\linewidth]{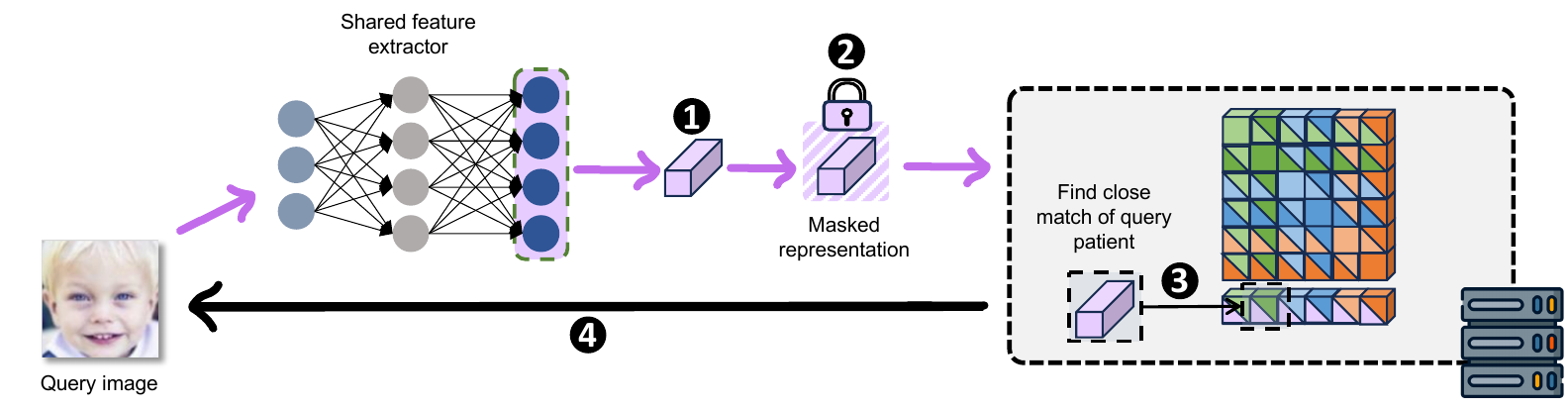}
    \caption{\textbf{(1)} Similar to the cosine distance matrix computation, the silo owning the query patient extracts the features of its query patient's image using the final aggregated model. \textbf{(2)} It masks the resulting latent representation and sends it to the aggregator. \textbf{(3)} The aggregator computes the cosine distance of the patient to other patients in the gallery from the masked latent representations. \textbf{(4)} Then, the aggregator responds the query patient using the syndromes of the gallery patients.}
    \label{fig:inference_on_query_patient}
\end{figure}

Once the cosine distance matrix and patient syndromes are available in the aggregator, the data owners can query about the syndromes of their patients privately, allowing them to benefit from the inference or diagnosis service while protecting patient privacy. To use these services, data owners first compute the latent representation of the query patient's image, then process this latent representation via FLAKE, and finally send it to the aggregator so that the aggregator computes the distance of the query patient's latent representation to other patients and responds in a determined manner such as returning the closest patient's syndrome. Figure \ref{fig:inference_on_query_patient} illustrates the described process of performing inference on the query patient. This inference is also possible for new data owners, joining the system later, after receiving the ensemble feature extractor model and the common seed for random generation in FLAKE. In addition to syndrome diagnosis, the aggregator could notify patient owners of exceptionally similar cases based on the cosine distance matrix, allowing for further studies to investigate these patients for the discovery of a new syndrome. 

\subsection{Dataset} \label{sec:dataset}
To preprocess the facial images of patients, we followed the same steps outlined by \citet{hustinx2023improving}. First, we cropped the images to ensure that only the patients' faces remained in the frame. Next, we rotated and aligned the cropped facial images to centralize the face within each image. Another crucial preprocessing step was to classify disorders based on their representation in our dataset. We note that all disorders in this study are considered rare ($<1:2000$) or even ultra-rare ($< 1:50,000$) in clinical terms due to their low incidence. However, for the purpose of our analysis, we use the label data-scarce or rare for syndromes with fewer than seven images and data-rich or frequent for those with seven or more. This approach highlights the influence of data availability on performance, which is logically distinct from how medically ‘rare’ a syndrome is.

\begin{figure}[!t]
    \centering
    \includegraphics[width=\linewidth]{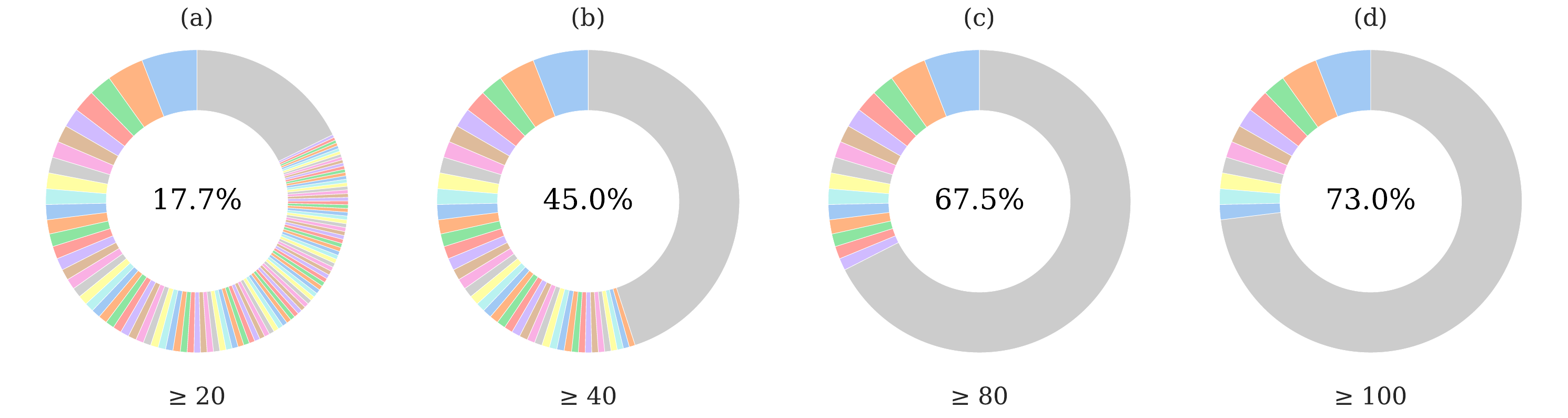}
    \caption{Distribution of class frequencies in the GMDB dataset for frequent images across four thresholds. Each panel shows a pie plot where slices represent classes occurring at or above the specified frequency threshold, and the grey slice aggregates all remaining classes below the threshold. The percentage in the center indicates the fraction of the dataset belonging to classes with fewer than the threshold number of samples. Panels correspond to thresholds of \textbf{(a)} $\geq20$, \textbf{(b)} $\geq40$, \textbf{(c)} $\geq80$, and \textbf{(d)} $\geq100$ samples.}
    \label{fig:gmdb_class_distribution}
\end{figure}

To give further insight about the dataset, Figure~\ref{fig:gmdb_class_distribution} shows the frequency distribution of frequent syndromes across four thresholds. For each threshold, syndromes occurring at or above the cutoff are shown individually, while the grey slice aggregates all syndromes with fewer samples. The percentage in the center denotes the fraction of the dataset belonging to these lower-frequency syndromes. Even among the frequent set, sample availability is highly limited where 17.7\% of the dataset consists of syndromes with fewer than 20 images, increasing to 45\% and 67.5\% for thresholds of 40 and 80 images, respectively. Only 11 syndromes have more than 100 images, together accounting for just 27\% of the data, with the remaining 73\% corresponding to 193 syndromes.

For a fair comparison to the centralized version, we used the same sets of training, validation, and test samples for both data-scarce (rare) and -rich (frequent) syndromes.
To simulate how hospitals (called silos) might collaborate in the system, we divided the dataset of frequent syndromes into 4, 8, and 16 groups. Each group represents one silo. We tried two ways of splitting the syndromes: 

Near-uniform distribution: We aimed to distribute patients from each syndrome evenly across all silos. For this scenario each silo would have roughly the same number of patients for each syndrome. This could be expected for most dominant disorders that occur at the same incidence independent of location. However, perfect balance was hard to achieve because some syndromes had very few samples. In the training data, each silo ended up with 194 syndromes, which is close to the full set of 204 syndromes. But for validation and testing, the distribution was less even, ranging from 155 syndromes in some silos to as few as 22 in others.

Extreme distribution: We created a scenario where each silo had completely different syndromes (no overlap). This might occur for some recessive disorders due to founder effects and it was simulated only for frequent syndromes and with 8 silos. This setup is much harder for the federated system because it can cause the local models in silos to become too different, making it harder for the global model to work well.

For data-scare syndromes, we split the data into 4, 8, and 16 groups more casually, as they were not used for training. The uneven distribution did not affect the model because rare syndromes are only used for testing and evaluation, not training.

In summary, we tested how different ways of splitting data (balanced vs. extreme) affected the system’s ability to train and predict accurately. The extreme scenario helped us understand how well the system could handle very challenging, uneven data setups.



\subsection{Experimental Setup} \label{sec:exp_setup}
To address the critical challenge of diagnosing rare syndromes with facial dysmorphism in a privacy preserving manner, we developed and tested our federated GestaltMatcher service through a series of experiments. Our experiments were conducted on a Linux server running Ubuntu 22.04, equipped with an Intel XEON CPU E5-2650 v4 with 256 GB of memory and GeForce GTX 1080 Ti GPUs with 11 GB of memory each.

In our experiments, we set the total number of epochs to 50, consistent with \citet{hustinx2023improving}. With the number of silos fixed at 8 and aggregating local models after each epoch, we compared the performance of the centralized and federated GestaltMatcher services. This experiment aimed to validate that the federated GestaltMatcher service provides similar performance to the centralized version. Additionally, for the 8-silo setup, we experimented with different aggregation intervals, selecting values from the set \{1, 5, 10, 25, 50\}, corresponding to 50, 10, 5, 2, and 1 aggregation(s), respectively. After the designated number of epochs, the local models were aggregated by the central aggregator and then redistributed to the silos. This setup allowed us to evaluate the impact of the aggregation interval on the performance of the federated GestaltMatcher service. Furthermore, we experimented with varying numbers of silos to assess the robustness of the federated GestaltMatcher service as data distribution varied. By fixing the aggregation interval of local models to 1 epoch, we analyzed the performance of the federated GestaltMatcher service for 4, 8, and 16 silos, respectively. Finally, to systematically evaluate robustness under heterogeneous client distributions, we simulated non-IID data partitions using a Dirichlet distribution with concentration parameter $\alpha \in \{0.5, 1, 5, 10\}$. The smaller $\alpha$ values correspond to more skewed client distributions, whereas the larger $\alpha$ values produce more balanced splits. This setup allowed us to analyze the effect of data heterogeneity on both predictive performance and representation-level stability of the federated GestaltMatcher service.

Throughout the experiments, performance evaluation was conducted following the approach of \citet{hustinx2023improving}. Specifically, if the syndrome of a query patient was among the Top-k unique closest syndromes, it was considered a Top-k match, where $k \in \{1, 5, 10, 30\}$. We also assessed the federated GestaltMatcher service's performance on different combinations of frequent and rare syndromes in the gallery and test sets where gallery set acts as reference set in the inference phase. The experiments included scenarios where: (1) only frequent syndromes were present in both test and training sets (F-F), (2) only rare syndromes were present in both test set and gallery (R-R), (3) frequent syndromes were present in both sets with rare syndromes included in the gallery set (F-FR), and (4) rare syndromes were present in both sets with frequent syndromes included in the gallery set (R-FR). This comprehensive evaluation framework enabled us to thoroughly assess the efficacy of the federated approach in various realistic settings.

\subsection{Experiments} \label{sec:exp}
In this section, we begin by comparing the performance of the federated GestaltMatcher service to the centralized GestaltMatcher service, which serves as our baseline, for both distributions. Next, we investigate the impact of the divergence of local models due to different aggregation intervals on the performance of the federated service. Finally, we assess the robustness of our federated GestaltMatcher service concerning the number of silos in the system.


\begin{figure}[!ht]
    \centering
    \begin{subfigure}[b]{0.47\linewidth}
        \centering
        \includegraphics[width=\linewidth]{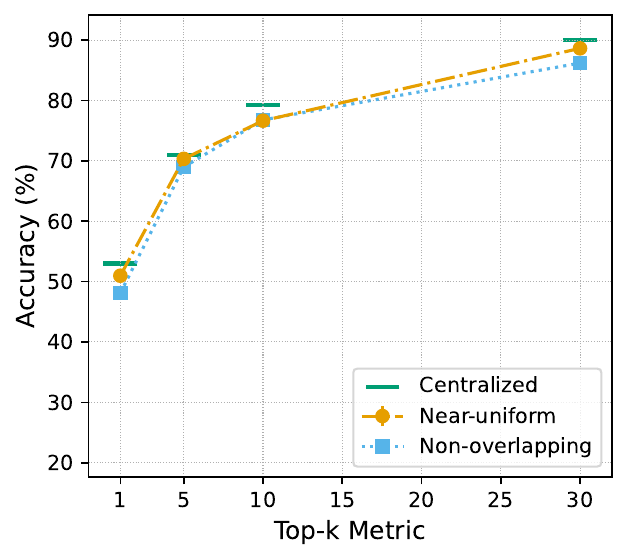}
        \caption{F-F}
        \label{fig:result_ff}
    \end{subfigure}
    \hfill
    \begin{subfigure}[b]{0.47\linewidth}
        \centering
        \includegraphics[width=\linewidth]{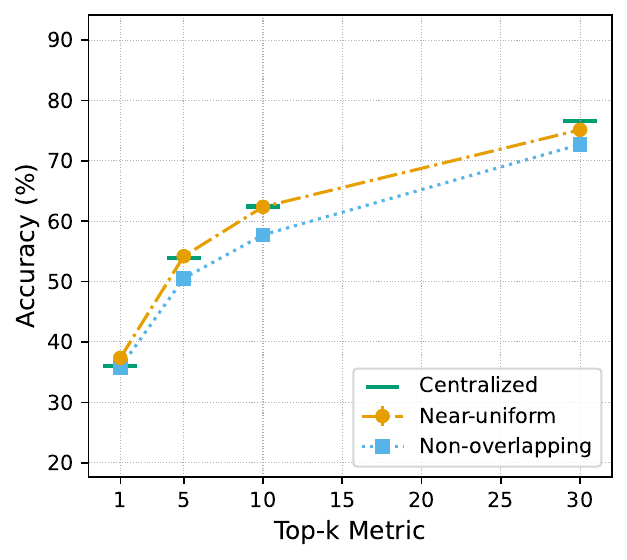}
        \caption{R-R}
        \label{fig:result_rr}
    \end{subfigure}\\
    \begin{subfigure}[b]{0.47\linewidth}
        \centering
        \includegraphics[width=\linewidth]{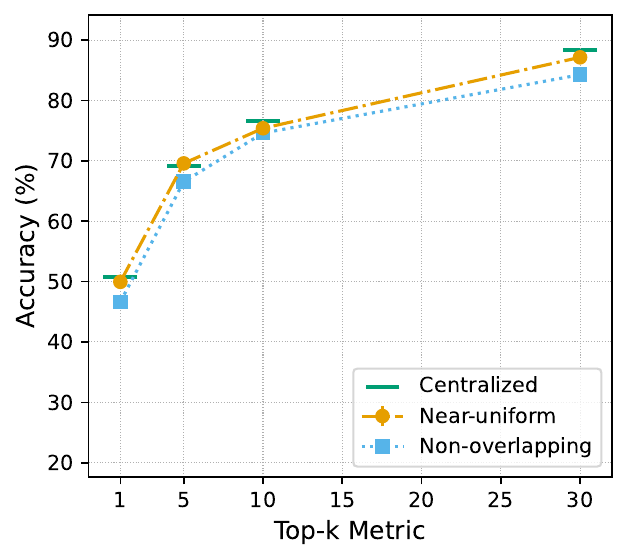}
        \caption{F-FR}
        \label{fig:result_ffr}
    \end{subfigure}
    \hfill
    \begin{subfigure}[b]{0.47\linewidth}
        \centering
        \includegraphics[width=\linewidth]{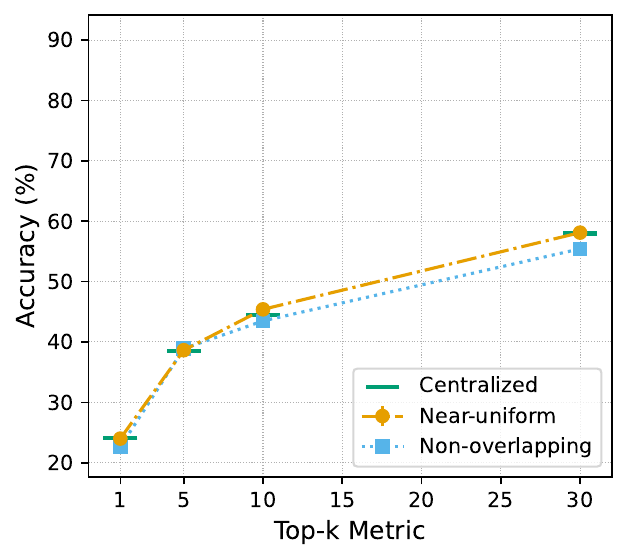}
        \caption{R-FR}
        \label{fig:result_rfr}
    \end{subfigure}
    \caption{The results of experiments with $8$ silos for both near-uniform and non-overlapping distributions of syndromes and their comparisons to the centralized version are demonstrated. The models' performance is evaluated using Top-1, Top-5, Top-10, and Top-30 matching accuracy in different test-gallery settings, that is frequent-frequent (F-F), rare-rare (R-R), frequent-frequent+rare (F-FR), and rare-frequent+rare (R-FR).}
    \label{fig:8_clients_results_uniform_and_nonoverlapping}
\end{figure}

\subsubsection{Federated vs Centralized}
We conducted the first set of experiments to compare the performance of the federated GestaltMatcher service with the centralized version. In these experiments, we set the number of participants to 8 and the aggregation interval of local models to 1 epoch, meaning the local models are aggregated after each local epoch by the aggregator. Using the Top-k matching accuracy, we compared the models' performance on frequent-frequent (F-F), rare-rare (R-R), frequent-frequent+rare (F-FR), and rare-frequent+rare (R-FR) settings, where the set before the “-” indicates the test set and the one after indicates the gallery set(s).

We compared both near-uniform and non-overlapping label distributions to the centralized version, as illustrated in Figure \ref{fig:8_clients_results_uniform_and_nonoverlapping}. The figure presents the average results from experiments conducted with five different near-uniform label distributions, demonstrating that the federated GestaltMatcher service achieves performance comparable to the centralized service across all test-gallery settings and Top-k metrics. Similarly, experiments with the non-overlapping label distribution yield competitive results, exhibiting only a slight performance decline compared to the near-uniform distribution.

When comparing results obtained with frequent syndromes in the test set to those with rare syndromes in the test set, we observe that the model performs significantly better for frequent syndromes. Among all settings, the R-FR scenario yields the lowest performance. It is important to note that this lower performance is not a consequence of using federated learning; the centralized version also performs poorly in this setting. This highlights the inherent challenges of accurately diagnosing rare syndromes, regardless of the model architecture used.

\begin{table}[!tb]
    \centering
    \begin{tabular}{cccccccc}
        \multicolumn{2}{c}{} & & \multicolumn{5}{c}{Aggregation interval (\# epochs)} \\
        \cmidrule{4-8}
         & & Metric & 1 & 5 & 10 & 25 & 50 \\
         
        \midrule
        \multirow{4}{*}{F-F} & \multirow{2}{*}{Near-Uniform} & Top-1 & \textbf{50.16} & 49.03 & 45.08 & 39.68 & 37.37 \\ \cmidrule(lr){3-8}
        & & Top-30 & 87.96 & \textbf{88.19} & 86.81 & 83.55 & 82.53 \\ \cmidrule(lr){2-8}
        & \multirow{2}{*}{Non-overlapping} & Top-1 & 48.10 & \textbf{48.37} & 46.41 & 41.39 & 37.66 \\ \cmidrule(lr){3-8}
        & & Top-30 & 86.19 & \textbf{87.37} & 86.08 & 85.93 & 82.80 \\ \cmidrule(lr){1-8}
        
        \multirow{4}{*}{R-R} & \multirow{2}{*}{Near-Uniform} & Top-1 & \textbf{37.25} & 37.09 & 36.37 & 34.52 & 35.42 \\ \cmidrule(lr){3-8}
        & & Top-30 & 74.72 & \textbf{76.55} & 75.78 & 75.27 & 75.17 \\ \cmidrule(lr){2-8}
        & \multirow{2}{*}{Non-overlapping} & Top-1 & 35.81 & 35.85 & 35.82 & 35.02 & \textbf{36.45} \\ \cmidrule(lr){3-8}
        & & Top-30 & 72.71 & 73.90 & 74.57 & 74.93 & \textbf{76.91} \\ \cmidrule(lr){1-8}

        \multirow{4}{*}{F-FR} & \multirow{2}{*}{Near-Uniform} & Top-1 & \textbf{49.16} & 48.22 & 44.08 & 38.57 & 36.25 \\ \cmidrule(lr){3-8}
        & & Top-30 & 86.33 & \textbf{87.65} & 84.44 & 82.07 & 80.21 \\ \cmidrule(lr){2-8}
        & \multirow{2}{*}{Non-overlapping} & Top-1 & 46.64 & \textbf{47.06} & 45.58 & 39.57 & 36.41 \\ \cmidrule(lr){3-8}
        & & Top-30 & 84.25 & \textbf{85.98} & 84.92 & 82.81 & 81.48 \\ \cmidrule(lr){1-8}

        \multirow{4}{*}{R-FR} & \multirow{2}{*}{Near-Uniform} & Top-1 & \textbf{23.59} & 23.11 & 22.84 & 21.21 & 21.65 \\ \cmidrule(lr){3-8}
        & & Top-30 & 57.60 & 57.70 & \textbf{58.35} & 55.80 & 54.89 \\ \cmidrule(lr){2-8}
        & \multirow{2}{*}{Non-overlapping} & Top-1 & 22.64 & \textbf{23.49} & 22.80 & 21.68 & 22.66 \\ \cmidrule(lr){3-8}
        & & Top-30 & \textbf{55.42} & 55.16 & 54.91 & 55.16 & 53.60 \\ 
        \bottomrule
    \end{tabular}
    \caption{Top-1 and Top-30 matching accuracies (\%) of models trained with different aggregation intervals, aggregating local models after a specified number of epochs, across various test-gallery settings for both near-uniform and non-overlapping distributions. The best performance for each metric and distribution is highlighted.}
    \label{tab:aggregation_period_results}
\end{table}

\subsubsection{Effect of Aggregation Intervals on Local Model Divergence}
In addition to comparing the federated GestaltMatcher service with the centralized version, we analyzed the impact of varying aggregation intervals on local model divergence. To understand how different aggregation intervals affect the performance of the federated service, we conducted experiments by fixing the number of silos to 8 and varying the aggregation intervals.

Table \ref{tab:aggregation_period_results} presents the results of these experiments, showing the Top-1 and Top-30 matching accuracies of models under different settings and distributions. For each test-gallery set, distribution, and metric, we highlighted the aggregation interval that yielded the best performance for the federated GestaltMatcher service. The results indicate that larger aggregation intervals, which allow for more divergence among local models, generally lead to decreased performance. Frequent aggregation helps maintain consistency among the local models, preventing them from drifting too far apart. Aggregating local models after every 5 epochs emerged as the best choice in 8 cases, making it the overall best-performing aggregation interval.

When examining the near-uniform distribution of the syndromes, aggregating local models after every epoch led to the best performance in all settings when evaluating with the Top-1 matching accuracy. For Top-30 matching accuracy in near-uniform distribution, the results favored an aggregation interval of 5 epochs. In contrast, for non-overlapping distributions, there was no consistent pattern across the settings. The results show that an aggregation interval of 5 epochs consistently led to slightly better performance when the test set comprised frequent syndromes. However, when rare syndromes were in the test set, there was no clear optimal aggregation interval.


\subsubsection{Evaluating Model Robustness Across Varying Silo Counts}
We further investigated how the number of silos impacts the performance of the federated GestaltMatcher service. Federated learning models may experience performance degradation as the number of silos increases, due to the potential reduction in data per silo and increased variability in the data distribution across silos. To assess the robustness of our federated GestaltMatcher service, we conducted experiments while keeping the total number of samples constant and varying the number of silos among 4, 8, and 16.

\begin{figure}[!t]
    \centering
    \includegraphics[width=0.9\linewidth]{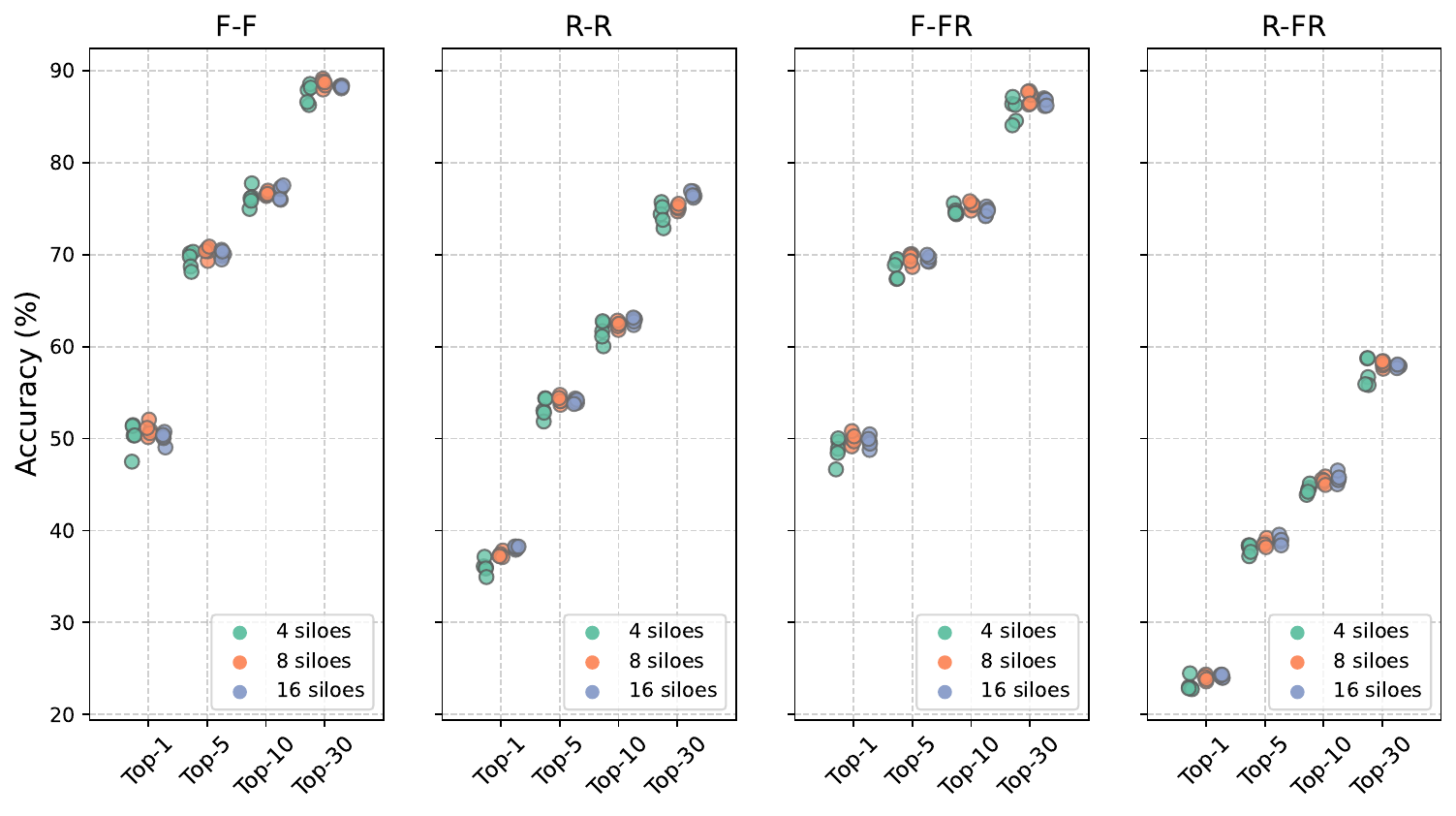}
    \caption{Results of experiments with near-uniform distribution, varying the number of silos across different test-gallery settings, with a fixed aggregation interval of 1 epoch. Each experiment was repeated 5 times, and model performance was evaluated using the Top-k metric. Individual experiment results are depicted separately to demonstrate the distribution.}
    \label{fig:num_of_silos}
\end{figure}

In these experiments, we set the aggregation interval of local models to 1 epoch and repeated each experiment five times to ensure statistical significance. The results were analyzed across different test-gallery settings, focusing on both frequent and rare syndromes. The performance was measured using the Top-k matching accuracies where $k \in \{1,5,10,30\}$. The results, summarized in Figure \ref{fig:num_of_silos}, indicate that the federated GestaltMatcher service maintains stable performance across different numbers of silos. The largest decrease in performance happened in rare-rare setting, particularly between the experiments with 4 and 16 silos, with a mean accuracy difference of $2.18$ in Top-30 matching accuracy. This indicates that the model's performance did not significantly deteriorate with an increasing number of silos, suggesting that the federated approach is robust to changes in silo count.

\subsubsection{Evaluating Model Robustness Across Varying Degrees of Data Heterogeneity}

An important challenge for federated learning in clinical contexts is the degree of data heterogeneity across participating sites. In addition to varying the number of silos, we analyzed the effect of non-IID data distributions on the performance of the federated GestaltMatcher service. To simulate different levels of heterogeneity across clients, we employed a Dirichlet distribution with concentration parameter $\alpha$. The smaller $\alpha$ values correspond to more skewed client distributions, whereas the larger $\alpha$ values produce more balanced splits.

\begin{figure}[!t]
    \centering
    \includegraphics[width=\linewidth]{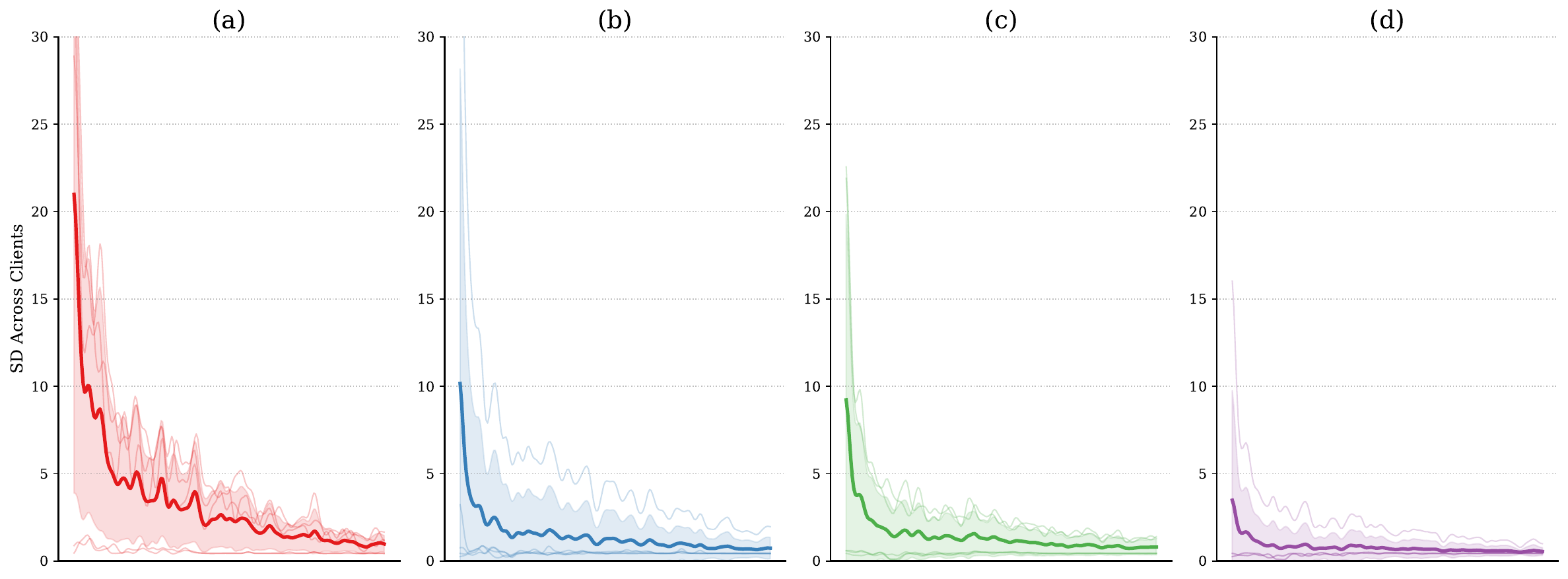}
    \caption{Standard deviation (SD) of class distributions across clients for different Dirichlet $\alpha$ values. The x-axis corresponds to individual classes, and the y-axis shows the SD of their proportions across clients. Each panel shows the SD per class for one $\alpha$, with semi-transparent lines representing individual distributions, and the bold line showing the mean SD with shaded area indicating $\pm$ standard deviation. The four panels correspond to $\alpha$ values of \textbf{(a)} $0.5$, \textbf{(b)} $1$, \textbf{(c)} $5$, and \textbf{(d)} $10$. Higher $\alpha$ values produce more balanced distributions across clients, whereas lower $\alpha$ values result in greater heterogeneity.}
    \label{fig:fed_dists}
\end{figure}

Figure \ref{fig:fed_dists} shows the standard deviation of class distributions across clients under different $\alpha$ values. As expected, lower $\alpha$ values (e.g., $\alpha=0.5$) result in highly heterogeneous splits, while higher values yield more uniform partitions. This setup allowed us to systematically evaluate the robustness of our system under varying degrees of data heterogeneity.

\begin{figure}[!t]
    \centering
    \includegraphics[width=0.9\linewidth]{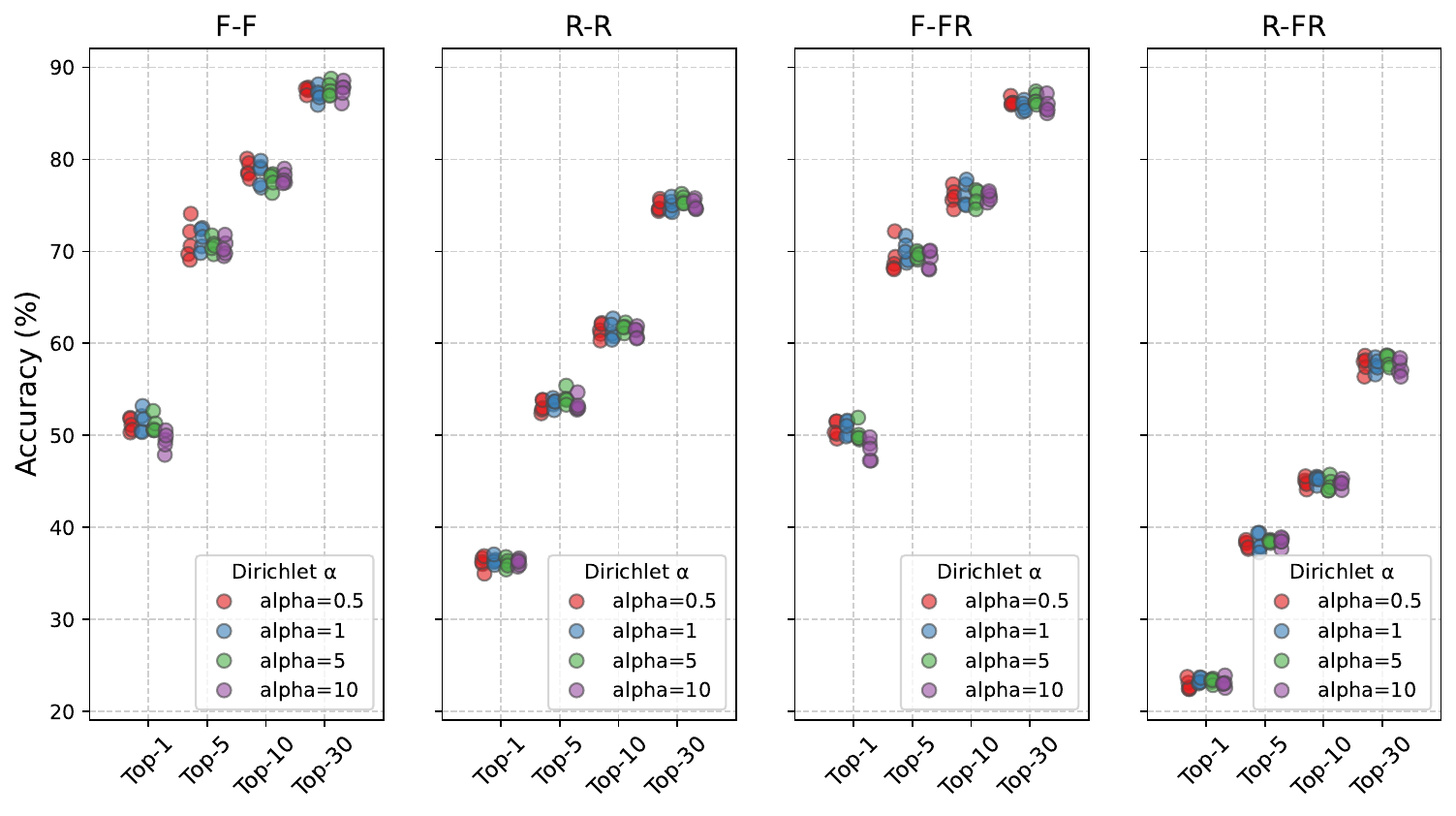}
    \caption{Results of experiments with 4 clients under varying Dirichlet distribution parameters $(\alpha = 0.5, 1, 5, 10)$ across different test-gallery settings, with a fixed aggregation interval of 1 epoch. Each experiment was repeated 5 times, and model performance was evaluated using the Top-k metric. Individual experiment results are shown separately to illustrate the variability across runs.}
    \label{fig:dirichlet_alphas}
\end{figure}

The results of these experiments, summarized in Figure \ref{fig:dirichlet_alphas}, show that the federated GestaltMatcher service maintained stable performance in all test–gallery settings and Top-k metrics, even in pronounced heterogeneity. These results demonstrate that the model provides robust performance regardless of data imbalance.

\begin{figure}[!t]
    \centering
    \includegraphics[width=\linewidth]{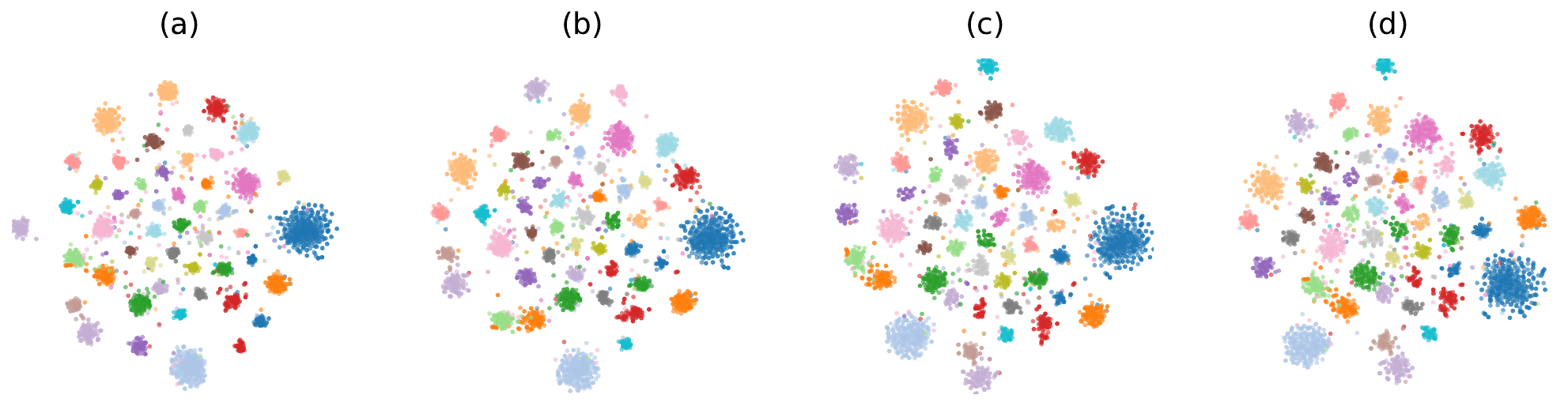}
    \caption{t-SNE visualization of the top 50 most frequent classes across four federated experiments with varying Dirichlet $\alpha$. Each panel shows the t-SNE embedding of the averaged feature representations for one experiment, colored by true class label. The four experiments correspond to $\alpha$ values of $0.5$ \textbf{(a)}, $1$ \textbf{(b)}, $5$ \textbf{(c)}, and $10$ \textbf{(d)}.}
    \label{fig:tsne_top50}
\end{figure}

To further investigate this robustness, we visualized the feature representations of the top 50 most frequent syndromes using t-SNE (Figure \ref{fig:tsne_top50}). For all Dirichlet parameters, the frequent syndromes formed well-separated clusters, indicating that the global feature extractor produces stable latent representations across different levels of data heterogeneity.


\begin{table}[ht]
\centering
\begin{tabular}{c|cc|cc}

\multirow{2}{*}{$\alpha$} & \multicolumn{2}{c|}{Intra-class} & \multicolumn{2}{c}{Inter-class} \\
 & Test & Rare & Test & Rare \\
\hline
0.5  & 0.775 $\pm$ 0.002 & 0.782 $\pm$ 0.006 & 0.965 $\pm$ 0.002 & 0.962 $\pm$ 0.002 \\
1.0  & 0.775 $\pm$ 0.009 & 0.777 $\pm$ 0.006 & 0.965 $\pm$ 0.003 & 0.962 $\pm$ 0.004 \\
5.0  & 0.764 $\pm$ 0.004 & 0.770 $\pm$ 0.006 & 0.967 $\pm$ 0.003 & 0.964 $\pm$ 0.003 \\
10.0 & 0.764 $\pm$ 0.006 & 0.767 $\pm$ 0.007 & 0.965 $\pm$ 0.003 & 0.962 $\pm$ 0.004 \\
\hline
\end{tabular}
\caption{Intra- and inter-class cosine distances (mean $\pm$ std) for test and rare gallery samples across five experimental runs at different Dirichlet $\alpha$ values.}

\label{tab:cluster_metrics}
\end{table}

To quantify representational stability under varying data heterogeneity, we computed intra- and inter-class cosine distances across test and rare gallery samples. Table~\ref{tab:cluster_metrics} reports the mean $\pm$ standard deviation over five experimental runs for different Dirichlet $\alpha$ values. Across all $\alpha$, the distances remain largely stable, with only minor variation between $\alpha = 0.5$ and $\alpha = 10$. Smaller intra-class distances and larger inter-class distances indicate that the global feature extractor preserves well-separated syndrome clusters, even under strongly non-IID conditions. These quantitative results support the robustness of the learned representations observed in the t-SNE visualizations (Figure~\ref{fig:dirichlet_alphas}).

\section{Discussion} \label{sec:discussion}
A decentralized healthcare data infrastructure reflects the reality of how healthcare systems are structured today. Hospitals and clinics naturally operate as independent entities, each managing its own data while adhering to privacy regulations. If medical artificial intelligences like GestaltMatcher are to be deployed within this structure while continuing to improve, they must be able to not only learn from but also perform syndrome diagnosis and discovery from distributed data. This means that instead of sharing raw data or its representation in plaintext, securely transformed summaries of image features are exchanged, which ensure privacy while retaining enough information for analysis.

To address these privacy concerns, we propose the federated GestaltMatcher service. This decentralized approach enables data owners to participate in collaborative efforts with improved privacy protection of their patient images. Our solution maintains the performance of the central model by outsourcing computation and inference tasks, which are syndrome diagnosis and discovery, to a server that only accesses masked latent representations of images. This method effectively prevents reconstruction or model inversion attacks, which could exploit the global model and latent image representations to reconstruct original images (see Section \ref{sec:image_rec} for more details). By ensuring that only masked encodings are shared, we uphold the privacy and security of patient data. Our experimental analyses demonstrated that the federated GestaltMatcher service provides the same utility as the centralized service that requires gathering all patient images in one place.

Additionally, we observed that the local ensemble feature extractor models benefit from relatively frequent aggregation for both distributions. Our results indicated that the best-performing models were typically trained with aggregation intervals of 5 epochs. We hypothesize that this frequency prevents the local models from diverging significantly while being trained locally. When comparing the impact of very frequent aggregation, such as after every local epoch, we found that aggregating after 5 epochs performed slightly better. This suggests that local models need more than a single epoch to learn meaningful patterns from the local data before being aggregated with others. Furthermore, the impact of aggregation intervals on model performance varied slightly depending on the test-gallery settings. For test patients with rare syndromes, there was no clear aggregation interval that consistently outperformed others, indicating that the lack of rare syndrome patients in the gallery might cause different responses to varying aggregation intervals in the test sets.

Our federated GestaltMatcher service is also designed to be robust against variations in the number of silos. We have observed no significant degradation in performance as the number of silos increases. This robustness ensures that the system can scale effectively, accommodating a growing number of participants without compromising accuracy. Moreover, the federated system is highly flexible in integrating new silos. If desired, the entire system can be retrained to incorporate data from new sources into the global encoder. Alternatively, new data can be integrated using the existing global ensemble feature extractor, making it immediately available for predictions and the discovery of ultra-rare syndromes. This flexibility is crucial for continuously updating and expanding the database to reflect new findings and emerging patient data.

Beyond scalability, an important strength of our approach is robustness to heterogeneous data distributions across silos. In practice, federated cohorts are rarely balanced, as institutions differ in sample size, syndrome prevalence, and population demographics. Our experiments show that the federated GestaltMatcher maintains stable performance even under strongly skewed non-IID partitions. Both the t-SNE visualizations and the quantitative cluster metrics confirm that frequent syndromes remain well separated in the latent space and that inter- and intra-class distances are consistent across various degrees of heterogeneity. These results indicate that our global feature extractor produces stable representations despite substantial distributional skew, which is essential for deployment in clinical settings where heterogeneity is unavoidable.

Given the stringent privacy regulations and the inherent sensitivity of patient data, our federated GestaltMatcher service holds significant potential to advance research on syndromes with facial dysmorphism. Individual data sources often lack sufficient data to train robust models independently. By supporting collaboration among multiple data owners while addressing their privacy concerns, our solution has a substantial real-life impact. Facilitating secure ensemble feature extractor model training for building a distributed database along with ensuring secure diagnosis and discovery within it can enhance diagnostic accuracy, improve patient outcomes, and enable the identification of previously unrecognized syndromes. The federated approach ensures that valuable data, which would otherwise remain siloed, can contribute to global medical research efforts.

\subsection{Limitations and Future Work} \label{sec:lim_and_fw}
In this study, we consider the security model of a semi-honest adversary. This assumption implies that while participants in the federated system will follow the protocol correctly, they may attempt to deduce additional information from the received data. Although this model provides a reasonable balance between security and practicality, it does not cover potential active adversaries who might deviate from the protocol to compromise the system. Another limitation is the potential need for verification of new data sources before they are allowed to join the system. This step may be necessary to ensure the integrity and reliability of the federated network. Verification processes could involve assessing the credibility of the institutions and the quality of their data. While this adds an extra layer of security, it also introduces additional overhead and complexity. Despite these limitations, they are often acceptable in the context of collaborations among large, reputable institutions. These institutions are typically semi-honest and can be verified by relevant authorities, ensuring a trustworthy network of data contributors.

For future work, several avenues could be explored to enhance the federated GestaltMatcher service. One potential direction is improving the robustness and security of the system against active adversaries. This could involve developing more advanced cryptographic techniques or implementing differential privacy measures to protect against potential threats. Additionally, exploring adaptive aggregation strategies that dynamically adjust based on the characteristics of the data and the network conditions could further enhance the performance of the federated system. Investigating ways to incorporate federated learning with other data modalities, such as genomic data or clinical records, could also expand the applicability and effectiveness of the federated GestaltMatcher service. By integrating a multi-modal approach in a privacy preserving manner, the system could provide deeper insights and more accurate diagnoses of dysmorphic syndromes.

By addressing these limitations and exploring these future directions, the federated GestaltMatcher service can be further refined and expanded, ultimately contributing to more robust and comprehensive research in the field of rare syndromes.

\section{Methods} \label{sec:methods}
In this section, we describe the methods that enable a privacy preserving federated GestaltMatcher service. This involves collaboratively training a global ensemble feature extractor and inferring patients' syndromes based on their images in the latent space without exposing the latent representations. We begin by detailing the collaborative training process of the global ensemble feature extractor. Next, we explain the computation of distances between patients' latent representations, which forms the basis of the collaborative inference task. Finally, we discuss how existing or new data sources can leverage the federated GestaltMatcher service for predicting syndromes in new patients or discovering exceptionally similar subgroups that may indicate a potential new syndrome.

\subsection{Federated Ensemble Feature Extractor Training} \label{sec:fed_encoder}
To train a global ensemble feature extractor using data distributed across multiple silos without revealing or centralizing the data, we employ federated learning. Each silo begins with a common local model, and it is crucial for the silos to agree on the number of output neurons. In many real-world scenarios, silos might not have patients with certain syndromes, making it difficult to determine the output layer size. To address this issue, the silos collectively determine the total number of different syndromes in the system at the start.

Once the common model is set up, each silo trains this model on its local data for a specified number of epochs. Following local training, we use federated averaging \cite{mcmahan2017communication} to aggregate the local model parameters at a central aggregator. To prevent information leakage through adversarial attacks on the local or aggregated models, we implement non-zero-sum masking, an adaptation of the zero-sum mask \cite{bonawitz2016practical}, designed to protect local data.

The zero-sum mask provides protection for local models in the aggregator against membership inference \cite{shokri2017membership} and model inversion \cite{fredrikson2014privacy,zhang2020secret} attacks. However, this leaves the aggregated model vulnerable to such attacks by the adversarial aggregator. By applying an additional mask via non-zero-sum masking, the aggregated model remains concealed from the aggregator, enhancing security. The data owners later remove this mask once the aggregator sends the masked aggregated model back to them.

In detail, $\numparties$ silos generate the same $\numparties$ random sets of values using a common shared seed. Each silo, except for the last one, adds its corresponding random value sets to its model parameters. The last silo subtracts the sum of the previously added random values from its model parameters but adds the last random value set, thereby hiding the aggregated model from the aggregator. When the aggregator sums the local models and computes their average, it only sees the masked version of the global model. After receiving the masked aggregated model, the silos subtract the division result of the last mask by the number of silos from the masked global model to obtain its plaintext form.

This cycle of local training and aggregation is repeated for a specified number of federated rounds. In the end, the silos obtain the final global ensemble feature extractor model that encodes the patients' facial images, making them ready for the computation of distances among patients' latent representations in a distributed setting.

\subsection{Secure Computation of Patient Cosine Distance} \label{sec:patient_sim}
After each silo computes the latent representation of their patients' facial images locally, they need to calculate the cosine distances between each pair of these encodings in the latent space. A naive federated approach would involve sharing these latent representations among the silos, enabling each silo to compute the distance matrix for inference purposes. However, this approach poses significant privacy risks, as sharing them can reveal sensitive patient information. We discuss the privacy concerns associated with latent representation sharing in more detail in Section \ref{sec:security}.

To mitigate these privacy risks, we employ a privacy enhancing technology for the distance computation between patients' facial images. The core of cosine distance computation is the dot product of the latent representations of the images. We utilize FLAKE \cite{hannemann2023privacy}, a specialized framework designed to compute the Gram matrix of feature vectors distributed across multiple sources while preserving privacy. FLAKE uses a unique masking scheme to ensure secure computation.

In brief, FLAKE allows sources to generate a common mask matrix using a shared random seed and a source-specific left inverse of this common matrix. The sources then mask their data matrices with these generated matrices. The masked data matrices are shared with a third party, which computes the Gram matrix. In its original application, FLAKE was used to train an SVM model over the Gram matrix. For our purposes, we adapt this framework to calculate the cosine distances between the latent representations of patients' facial images distributed across multiple silos.

Once the Gram matrix is computed at the aggregator without revealing the latent representations, the entries of this matrix can be used to calculate the cosine distances between all pairs of patients' facial images in the federated GestaltMatcher service using the following formula:

\begin{equation}
\begin{split}
    \cd(\pone, \ptwo) &= 1 - \frac{\sum_{i=1}^{f} \pone_i \ptwo_i}{\sqrt{\sum_{i=1}^{f} \pone_i^2} \sqrt{\sum_{i=1}^{f} \ptwo_i^2}} \\
    &= 1 - \frac{G(\pone, \ptwo)}{\sqrt{G(\pone, \pone)} \sqrt{G(\ptwo, \ptwo)}}
\end{split}
\end{equation}

where $\cd(.,.)$ denotes the cosine distance function, $\pone$ and $\ptwo$ are the latent representations of patients' facial images computed using the global ensemble feature extractor, $\fea$ is the size of the latent space, and $G(.,.)$ is a function returning the corresponding dot product value of given feature vectors from the Gram matrix.

After computing the cosine distance matrix and sharing the syndromes of patients with the aggregator, the system is prepared to perform inference on patients and discovery of new syndromes.

\subsection{Inference Process and Novel Syndrome Detection} \label{sec:inference}
The federated GestaltMatcher service is capable of performing inference on patient facial images once the cosine distance matrix of gallery set patients and their syndromes is available on the aggregator. To evaluate the performance of the federated GestaltMatcher service on test images, the silos first compute the latent representations of their test images and then mask these representations using the same random masks employed for the gallery samples. This allows the aggregator to integrate these test samples with the existing gallery samples and compute the cosine distances between them. Once the cosine distance of a test sample to all gallery samples is computed, the prediction on the test sample can be made based on the chosen prediction strategy. In our experiments, we determine if the test sample's syndrome is among the closest Top-k unique syndromes in the gallery set, where $k \in \{1,5,10,30\}$. If the syndrome is among the Top-k closest, it is considered a correct prediction; otherwise, it is evaluated as a misclassification.

For patients from silos that did not participate in the global ensemble feature extractor training phase, the process is similar to those that did. New silos first obtain the global ensemble feature extractor and random seed used to generate masks for FLAKE. Then, they compute the latent representation of their patients' facial images and corresponding mask matrices. Thereafter, new silos follow the same steps as those involved in the training phase.

One of the major advantages of GestaltMatcher is its ability to discover new syndromes. In addition to predicting the syndrome of a query patient from the pool of existing syndromes in the system, GestaltMatcher allows for the identification of exceptionally similar small patient groups that could represent an undiscovered ultra-rare syndrome. Such patient groups can be identified using the cosine distance matrix, and further studies on these patients can be conducted. In the federated GestaltMatcher service, these discoveries are made in a privacy preserving manner, with only the silos having patients in the corresponding subgroup needing to arrange the necessary permissions and approvals. This approach saves the effort and resources of silos not having patients in these groups from being wasted.

\section{Security Analysis} \label{sec:security}


In this section, we evaluate the security of our federated GestaltMatcher service to ensure that it does not leak any information to any unauthorized party during computation. We start by presenting our security model, outlining the underlying assumptions and the adversarial capabilities we consider. Next, we evaluate the security and privacy aspects of training the global ensemble feature extractor model. Finally, we analyze the security measures in place for the cosine distance calculation and the inference on patients' facial images.

\subsection{Security Model and Assumptions}
In our security analysis, we use a semi-honest adversary model, also known as an honest-but-curious adversary. In this model, the corrupted parties follow the prescribed protocol, but attempt to infer sensitive information from the data they receive. We assume that these corrupted parties do not collude or collaborate in their efforts to extract sensitive information. This non-collusion assumption is critical to maintaining the integrity of the federated learning process.

At the beginning of the computation, we assume that the silos agree on a common seed to generate the same set of random values. They also agree on the model architecture and know the size of their frequent syndrome sets' union. This allows them to form the same initial local model at the beginning. Lastly, we assume that all communications between parties occur over secure channels, such as those protected by Transport Layer Security (TLS), to prevent data interception or tampering during transmission.

\subsection{Security During Global Ensemble Feature Extractor Model Training}
To train the global ensemble feature extractor model, the silos employ federated learning. However, federated learning alone does not guarantee complete security. For instance, the aggregator can perform membership inference attack on the global model when it is visible to the aggregator \cite{nasr2019comprehensive,hu2021source}. Besides, the local models sent in plaintext allow the aggregator to perform such attacks on the local model level, narrowing down the victim set to a single data source. To address these privacy concerns, we utilize secure aggregation. Specifically, we adapt the zero-sum mask secure aggregation to a non-zero-sum mask method. In this secure aggregation approach, neither the local models nor the aggregated global model are visible to the aggregator due to the masking applied to the global model. Once the masked global model is sent back to the data sources, they remove the remaining mask to obtain the plaintext global model. This process prevents the aggregator from performing membership inference attacks on both the global and local models.

\subsection{Facial Image Reconstruction from Latent Representations} \label{sec:image_rec}
\begin{figure}[!ht]
    \centering
    \begin{subfigure}{0.489\linewidth}
        \centering
        \includegraphics[width=\linewidth]{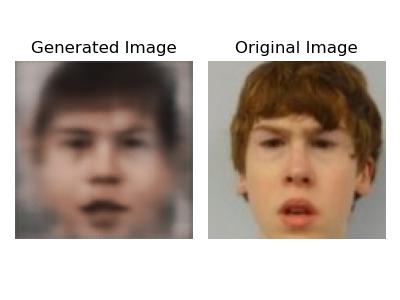}
        \caption{}
        \label{fig:rec_img_1}
    \end{subfigure}
    \hfill
    \begin{subfigure}{0.49\linewidth}
        \centering
        \includegraphics[width=\linewidth]{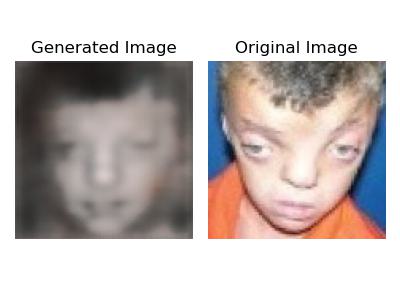}
        \caption{}
        \label{fig:rec_img_2}
    \end{subfigure}
    \caption{The original image of a patient and the reconstructed image of this patient using the corresponding encoding. \textbf{(a)} The reconstructed images visually have a sufficient resemblance to the original images. \textbf{(b)} This is especially true if the phenotype of a patient is highly unique.}
    \label{fig:reconstructed_patient_img}
\end{figure}

The naive federated approach for silos to proceed with cosine distance calculation and inference after training the global ensemble feature extractor model is to share the latent representations of their patients. However, this method poses significant privacy risks. To demonstrate this, we train a decoder or image reconstruction network using the training and validation samples. We then reconstruct the patients' facial images in the test set using only their latent representations. This mimics a scenario where silos share the latent representations of their patients' facial images, and one of them has enough samples to train a decoder on the global ensemble feature extractor model. Our experiments showed that it is possible to reconstruct the original images to a certain extent, revealing non-designated information such as gender. In some extreme cases, the uniqueness of a patient's phenotype could allow complete identification. For example, Figure \ref{fig:reconstructed_patient_img} shows the original and reconstructed images of a patient. Figure \ref{fig:rec_img_1} demonstrates a better reconstruction, highly resembling the original image based on visual inspection, whereas Figure \ref{fig:rec_img_2} illustrates a case where, despite imperfect reconstruction, the unique phenotype of a patient could reveal their identity.

Since the scope of this paper is not to perform a model inversion attack on the global ensemble feature extractor model, we do not further optimize the model to improve the performance of the attack. Instead, we demonstrate that sharing latent representations among silos is neither secure nor private, justifying the use of FLAKE. To further improve the performance of model inversion attacks, more sophisticated approaches can be employed \cite{zhang2020secret,wang2021variational}.

To mitigate these privacy concerns, we employ FLAKE to compute the cosine distance matrix. Using a common mask matrix among the silos and their individual left inverses allows them to outsource the computation of the Gram matrix to the aggregator, from which the cosine distance matrix will be computed. Since there is no information exchange among the silos, it is impossible for a silo to train a decoder to reconstruct the original image of a patient from another silo. As the silos only share their masked encodings with the aggregator, the aggregator cannot infer anything about the patients' facial images or their latent representations. The aggregator only obtains the Gram matrix of the latent representations of patients' facial images, which is the desired output for the silos. Since the labels are not considered private and cannot be used to deduce information about the latent representations, the silos share them with the aggregator in plaintext. Thus, our federated GestaltMatcher service enables privacy preserving cosine distance computation and inference on patients' facial images. For a detailed security analysis of using FLAKE, we refer the reader to the corresponding paper \cite{hannemann2023privacy}.

\section{Data Availability}
The dataset used in this study can be accessed upon request at \href{https://db.gestaltmatcher.org/documents}{https://db.gestaltmatcher.org/documents}. We followed the preprocessing steps described by \citet{hustinx2023improving} to prepare the patient images. A detailed description of these preprocessing steps is also available in the GitHub repository of the federated GestaltMatcher service.

\section{Code Availability}
The source code for the federated GestaltMatcher service is available at \href{https://github.com/mdppml/Federated_GestaltMatcher_Service}{https://github.com/mdppml/Federated\_GestaltMatcher\_Service}. The code is implemented in Python and includes the necessary documentation to replicate the study. The repository provides detailed instructions and sample scripts for setting up and running the experiments.

\section*{Acknowledgments}
This study is supported by the German Ministry of Research and Education (BMBF), project number 01ZZ2010.

\begin{appendices}



\section{Facial Image Reconstruction}
To reconstruct the facial images of patients from their latent representations, we designed a model utilizing 2D transposed convolutional layers implemented in PyTorch. The architecture is summarized in Table \ref{tab:img_rec_architecture}.

The model begins with a fully connected layer that transforms the 512-dimensional latent vector into a $7 \times 7 \times 256$ feature map. This serves as the starting point for a series of transposed convolutional layers, progressively increasing the spatial dimensions of the feature maps while reducing the number of channels. All \textit{ConvTranspose2d} layers, except the final one, are followed by a batch normalization layer (\textit{BatchNorm2d}) and a rectified linear unit (ReLU) activation function (\textit{ReLU}). These layers ensure that intermediate feature maps maintain numerical stability and non-linearity, aiding the model's ability to learn complex features.

In the final layer, after the last \textit{ConvTranspose2d} operation, we apply a hyperbolic tangent activation function (\textit{tanh}). This choice aligns with the preprocessing of input images, where pixel values are normalized between $-1$ and $1$. The \textit{tanh} activation ensures the output pixel values fall within this range, making it suitable for the reconstruction of normalized images.

The detailed architecture of the model, including the kernel size, stride, padding, and output dimensions at each stage, is provided in Table \ref{tab:img_rec_architecture}.

\begin{table}[!ht]
    \centering
    \caption{Model architecture to reconstruct patient facial images}
    \begin{tabular}{@{}ccccc@{}}
        \toprule
        \textbf{Layer} & \textbf{Kernel Size} & \textbf{Stride} & \textbf{Padding} & \textbf{Output Size (H $\times$ W $\times$ C)} \\ \midrule
        Fully Connected (Input: 512) & --         & --     & --     & $7 \times 7 \times 256$ \\ \midrule
        ConvTranspose2d & $4 \times 4$ & 2      & 1      & $14 \times 14 \times 128$ \\
        ConvTranspose2d & $4 \times 4$ & 2      & 1      & $28 \times 28 \times 64$ \\
        ConvTranspose2d & $4 \times 4$ & 2      & 1      & $56 \times 56 \times 32$ \\
        ConvTranspose2d & $4 \times 4$ & 2      & 1      & $112 \times 112 \times 16$ \\
        ConvTranspose2d & $3 \times 3$ & 1      & 1      & $112 \times 112 \times 8$ \\
        ConvTranspose2d & $3 \times 3$ & 1      & 1      & $112 \times 112 \times 3$ \\ \bottomrule
    \end{tabular}
    \label{tab:img_rec_architecture}
\end{table}

\end{appendices}


\bibliography{sn-bibliography}

\end{document}